\documentclass[journal]{IEEEtran}
\usepackage{amsmath,amsfonts}
\usepackage{array}
\usepackage{textcomp}
\usepackage{stfloats}
\usepackage{url}
\usepackage[]{footmisc}
\usepackage{amssymb}
\usepackage{multirow}
\usepackage{algorithm}
\usepackage{algpseudocode}
\usepackage{amsthm}
\usepackage{verbatim}
\usepackage{graphicx}
\usepackage{soul}
\usepackage{xcolor}
\usepackage{multicol}
\usepackage{wrapfig}
\usepackage{subcaption}
\usepackage[none]{hyphenat}
\usepackage{epstopdf}
\usepackage[]{color-edits}
\usepackage{cite}
\usepackage{textcomp}
\usepackage{fancyhdr}
\usepackage{booktabs}
\usepackage{listings}
\usepackage{mdframed}
\usepackage[normalem]{ulem}

\colorlet{punct}{red!60!black}
\definecolor{background}{HTML}{FFFFFF}
\definecolor{delim}{RGB}{20,105,176}
\colorlet{numb}{magenta!60!black}

\lstdefinelanguage{json}{
    basicstyle=\normalfont\ttfamily,
    stepnumber=1,
    numbersep=8pt,
    showstringspaces=false,
    breaklines=true,
    backgroundcolor=\color{background},
    literate=
     *{0}{{{\color{numb}0}}}{1}
      {1}{{{\color{numb}1}}}{1}
      {2}{{{\color{numb}2}}}{1}
      {3}{{{\color{numb}3}}}{1}
      {4}{{{\color{numb}4}}}{1}
      {5}{{{\color{numb}5}}}{1}
      {6}{{{\color{numb}6}}}{1}
      {7}{{{\color{numb}7}}}{1}
      {8}{{{\color{numb}8}}}{1}
      {9}{{{\color{numb}9}}}{1}
      {:}{{{\color{punct}{:}}}}{1}
      {,}{{{\color{punct}{,}}}}{1}
      {\{}{{{\color{delim}{\{}}}}{1}
      {\}}{{{\color{delim}{\}}}}}{1}
      {[}{{{\color{delim}{[}}}}{1}
      {]}{{{\color{delim}{]}}}}{1},
}

\def\BibTeX{{\rm B\kern-.05em{\sc i\kern-.025em b}\kern-.08em
    T\kern-.1667em\lower.7ex\hbox{E}\kern-.125emX}}
\usepackage{balance}

\usepackage{algorithm}

\newcommand{{\ours}}{Q-FedPD}

\begin{document}
\title{A Web-Based Solution for Federated Learning with LLM-Based Automation}
\author{\IEEEauthorblockN{Chamith Mawela, Chaouki Ben Issaid, and Mehdi Bennis}\\
\thanks{
C. Mawela, C. Ben Issaid, and M. Bennis are with the Centre of Wireless Communications, University of Oulu, 90014 Oulu, Finland. Email: \{chamith.mawelamudiyanselage, chaouki.benissaid, mehdi.bennis\}@oulu.fi}
}

\maketitle

\begin{abstract}
Federated Learning (FL) offers a promising approach for collaborative machine learning across distributed devices. However, its adoption is hindered by the complexity of building reliable communication architectures and the need for expertise in both machine learning and network programming. This paper presents a comprehensive solution that simplifies the orchestration of FL tasks while integrating intent-based automation. We develop a user-friendly web application supporting the federated averaging (FedAvg) algorithm, enabling users to configure parameters through an intuitive interface. The backend solution efficiently manages communication between the parameter server and edge nodes. We also implement model compression and scheduling algorithms to optimize FL performance. Furthermore, we explore intent-based automation in FL using a fine-tuned Language Model (LLM) trained on a tailored dataset, allowing users to conduct FL tasks using high-level prompts. We observe that the LLM-based automated solution achieves comparable test accuracy to the standard web-based solution while reducing transferred bytes by up to 64\% and CPU time by up to 46\% for  FL tasks.  Also, we leverage the neural architecture search (NAS) and hyperparameter optimization (HPO) using LLM to improve the performance. We observe that by using this approach test accuracy can be improved by 10-20\% for the carried out FL tasks.  
\end{abstract}


\section{Introduction}  
The recent advancements in machine learning (ML) have led to a surge in organizations adopting these technologies in their business solutions, driven by the abundance of user-generated data. However, the conventional centralized approach to model training poses a threat to user privacy, as raw client data must be shared with an external party. Federated learning (FL) \cite{mcmahan2017communication} has emerged as a novel paradigm that enables decentralized model training without sharing raw data. FL addresses this issue by enabling collaborative learning where only model weights are communicated among edge nodes and a parameter server (PS), minimizing the risk of private client data leakage. Since its inception by Google\cite{mcmahan2017communication}, FL has been adopted across various domains, such as global COVID diagnosis \cite{Xu2020.05.10.20096073}, vehicular networks \cite{9205482}, and wireless communications \cite{fedwire1, fedwire2, fedwire3}. 

Despite its potential, the adoption of FL is hindered by the complexity of building reliable communication architectures and the need for ML and network programming expertise. In fact, researchers and stakeholders participating in FL implementations must rely on programming expertise and develop seamless communication architectures between edge nodes and the PS. The FL community primarily relies on frameworks such as Flower \cite{flower}, PySyft \cite{PySyft}, and TensorFlow Federated \cite{TensorflowFL}, which require both programming and FL domain expertise. Flower and PySyft follow a similar design architecture where the PS and the client services are programmed separately and deployed into respective devices. Flower uses the remote procedure call (gRPC) protocol for communication between clients and PS while PySyft uses WebSockets.  TensorFlow Federated on the other hand is developed only for standalone simulation of FL setting on a single device.

In frameworks such as Flower and PySyft, the model architecture and hyperparameters for the FL task are typically hardcoded within the PS or client programs. This setup presents a significant challenge during iterative experimentation, as modifications to the model or its parameters require altering the code on either the PS or client devices. Consequently, only individuals with access to these devices can conduct iterative experiments on the FL process by adjusting the parameters. This motivates us to develop a web-based solution where the clients and PS serve as Application programmable interface (API) endpoints. This eliminates the requirement for direct access to any device to change parameters since it can be done using the API endpoint. Through the provided solution we eliminate the need for programming to carry out FL tasks since the solution will be provided as a web application. Another major drawback of the frameworks mentioned above is the lack of inherent support for model compression schemes. When the size of the model used for the FL process becomes bigger, the possibility of communication bottlenecks occurring could become higher in limited bandwidth communication settings. Hence to alleviate the potential bottlenecks in the FL communication process, our proposed solution will provide support for model compression schemes such as quantization \cite{quantization, quantization2} and sparsification \cite{sparse}. In Table \ref{tab:comparison}, we summarize the main difference between our framework and the existing solutions.

\begin{table*}[t]
\centering
\caption{Comparison of our proposed framework with existing solutions.}
\label{tab:comparison}
\resizebox{\textwidth}{!}{%
\begin{tabular}{@{}lcccc@{}}
\toprule
\textbf{Feature} & \textbf{Our Solution} & \textbf{Flower} & \textbf{PySyft} & \textbf{TensorFlow Federated} \\
\midrule
Web-based solution (No programming required) & \checkmark & $\times$ & $\times$ & $\times$ \\
\midrule
Built-in model compression & \checkmark & $\times$ & $\times$ & $\times$ \\
\midrule
Built-in scheduling algorithms & \checkmark & $\times$ & $\times$ & $\times$ \\
\midrule
Distributed implementation & \checkmark & \checkmark & \checkmark & $\times$ \\
\midrule
FL without direct device access & \checkmark & $\times$ & $\times$ & $\times$ \\
\midrule
Communication protocol & WebSockets & gRPC & WebSockets & N/A \\
\midrule
Intent-based automation of FL using LLM & \checkmark & $\times$ & $\times$ & $\times$ \\
\midrule
Automated NAS and HPO for FL & \checkmark & $\times$ & $\times$ & $\times$ \\
\bottomrule
\end{tabular}%
}
\end{table*}

More recently, due to the success of Large Language Models (LLMs) and their ability to fine-tune downstream tasks, a wide use of LLMs for automating various tasks can be observed. LLM-integrated tasks span from implementing co-pilots to fully automated solutions. Most of the solutions follow the same structure where an intent or instruction is given by a third party to the LLM where the intent holds necessary information for the LLM to proceed further to produce the intended output. Mostly the instruction is generated by a human or in some cases by the program itself. For instance, LLMs were used to solve math problems \cite{math}, automate real-world API consumption \cite{toolllm}, generate Verilog and Golang code \cite{code_generation_2, code_generation_1}, and automate IT infrastructure management using few-shot learning \cite{appleseed} or repair Ansible scripts for edge cloud infrastructure management \cite{ansible}. To the best of our knowledge, there is no existing research in the literature that addresses the automation of FL through user prompts. This study aims to fill this gap by exploring the potential of leveraging user prompts to automate the federated learning process, making it accessible for users with less FL domain expertise.

In FL tasks, the performance heavily depends on the architecture of the global model and the hyperparameters employed during training. In traditional ML contexts, neural architecture search (NAS) \cite{naspaper} and hyperparameter optimization (HPO) \cite{feurer2019hyperparameter} are commonly utilized to identify the optimal model architecture and hyperparameters tailored to a specific dataset. Hence, in our work, we investigate utilizing NAS and HPO to improve the performance of the LLM-based automated solution. Usually, NAS and HPO involve creating a search space for model architectures and hyperparameters, respectively, and carrying out an exhaustive search on the created search space to find the optimal configuration. Due to the dynamic nature of our automated solution and the capability of LLM in producing model architectures, we utilize LLM to create the search space for NAS.   

The main challenge in implementing FL tasks using multiple hardware entities as edge nodes and a PS is the need to build a reliable communication architecture from scratch. Our goal is to bridge the gap between conceptualizing and implementing an FL solution. We will address the limitation posed by other frameworks, by making the FL process accessible to parties without direct access to the devices running PS and clients. We also propose a solution to automate the FL process for user prompts, which we believe would make FL tasks more accessible to users with less domain knowledge. Accordingly, we introduce a web-based solution with an efficient communication architecture developed using WebSockets, to orchestrate FL tasks easily. Furthermore, we integrate an intent-based platform into the developed solution, that uses a fine-tuned LLM to carry out federated learning operations based on user prompts. Our contributions are as follows
\begin{itemize}
    \item We develop a standard web application supporting the FedAvg algorithm \cite{mcmahan2017communication} for performing FL tasks. A user-friendly web interface is developed so users can easily select the required parameters for the FL tasks. Furthermore, an efficient communication architecture based on WebSockets is developed as the back-end solution to handle communication between the PS and the edge nodes.
    \item We explore the implementation of secondary functionalities in FL, such as model compression to optimize the transmission of model weights and scheduling algorithms to manage client participation in the FL task.
    \item We investigate intent-based automation of the developed solution, where federated learning tasks are carried out according to a provided high-level prompt. To achieve this, we fine-tune an open-source LLM on a tailored dataset and integrate the fine-tuned LLM into the developed solution to automate the FL process based on user prompts.
    \item To further enhance the performance of the automated solution, we implement NAS and HPO. Specifically, we investigate the use of LLM to generate the search space for NAS and develop a selective halving-based approach for the HPO process to find the best model architecture and hyperparameters required for a specific FL task.
 
\end{itemize}
The rest of the paper is organized as follows. FL and related secondary functionalities are detailed in Section \ref{section2}. Section \ref{section3} gives a brief introduction to LLM and their fine-tuning process. The system design and the methodology of the proposed solution are discussed in Section \ref{section4}. An evaluation of the experimental results for the provided solution is discussed in Section \ref{section5}. Finally, the paper concludes with final remarks in Section \ref{section6}.

\section{Federated Learning}\label{section2}
\begin{figure}[t]
  \begin{center}
    \includegraphics[scale=0.26]{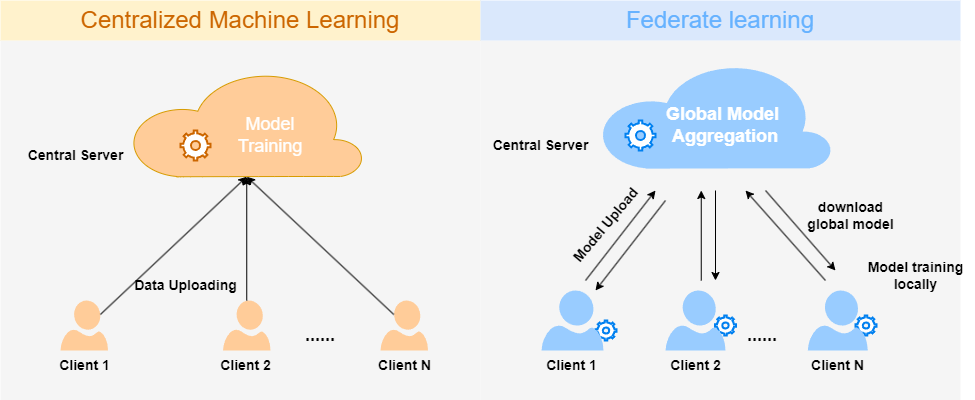}
  \end{center}
  \caption{Centralized Machine Learning vs Federated Learning.}
  \label{fig:centralvfed}
\end{figure}

FL is a paradigm of ML where models are trained collaboratively using multiple edge nodes (clients), and a PS. Unlike traditional centralized ML training, data is not communicated to a central server. Instead, model optimization occurs locally on the clients, preserving data privacy, as shown in Figure \ref{fig:centralvfed}. The objective of FL is to minimize the weighted average of the local objective functions of each client as follows
\begin{align}
    \min_{w\in \mathbb{R}^d} f(w) = \sum_{k=1}^{K} \frac{n_k}{n} F_k(w),
    \label{eq:fed_objective}
\end{align}
where $F_k(w)$ is the local objective function for the client $k$. It is assumed here that the total dataset is partitioned among $K$ number of clients where $n_k = |\mathcal{P}_k| $ and $\mathcal{P}_k$ is the set of data points assigned for the $k$th client. With these objectives in mind, Federated Averaging (FedAvg) has been proposed in \cite{mcmahan2017communication}. Considered one of the most widely used and influential algorithms in the realm of FL, FedAvg has inspired numerous extensions and improvements. In the remainder of this paper, we restrict ourselves to using FedAvg in our implementation. 

One of the major challenges in FL is scalability. As the model size and the number of clients grow, encountering a communication bottleneck in FL training may become inevitable. To alleviate the communication bottleneck related to the model size, model compression schemes have been adopted in the FL literature. Next, we review some common techniques for reducing the size of the models.
\subsection{Quantization}
Quantization is a technique for making models compact by reducing the precision of tensors. Usually, the $32$-bit precision of \textit{float32} is reduced to a lower precision, e.g., an 8-bit precision of \textit{int8}. A mapping between the floating point $x$ and its quantization $x_q$ is defined as
\begin{align}
\mathcal{Q}: x \in [\alpha, \beta] \rightarrow x_q \in [\alpha_q, \beta_q].
\end{align}
The quantization function $\mathcal{Q}(\cdot)$ \cite{quantization} is formulated using \textit{affine mapping} between floating point value and its quantized value, and using the \textit{clip} function as follows
\begin{align}
\mathcal{Q}(x,S,Z) = clip(round(\frac{x}{S} +Z), \alpha_q, \beta_q),
\label{eq:quantization}
\end{align}
where the clipping function is defined as
\begin{align}
clip(x,\alpha, \beta) =
\begin{cases}
\alpha & \text{if } x < \alpha \\
x & \text{if } \alpha \leq x \leq \beta \\
\beta & \text{if } x > \beta.
\end{cases}
\label{eq:clip}
\end{align}
Here, $S$ is an arbitrary positive real number, and $Z$ maps the 0 in floating value to the corresponding value in quantized data type. $S$ and $Z$ represent the scale and zero point quantization parameters, respectively. The recovery of the floating point value through de-quantization can be done using the following equation
\begin{align}
f_d(x_q, S, Z) = S(x_q - Z).
\end{align}
When applying quantization in the FL setting, locally trained models are quantized and communicated to the PS, where de-quantization is done to recover the original model weights before doing the model aggregation.
\subsection{Sparsification}
Sparsification \cite{sparse} is an alternative approach to model compression, which involves selectively retaining only a subset of the gradients through a sparse selection process. The most widely used sparsification methods in FL are $Top_k$ and $Rand_k$ where given a parameter $1 \leq k \leq d$ for a gradient vector $x \in \mathbb{R}^d$, only a $k$ subset of gradients selected will be communicated to the PS. 
\begin{itemize}
\item {$\mathbf{Top_k}$}: The top $k$ percent of gradients with the highest absolute values are selected while other values are set to zero. $Top_k: \mathbb{R}^d \rightarrow \mathbb{R}^d$ is formulated as below.

\begin{equation}
    [Top_k(x)]_j := \begin{cases}
        [x]_{\alpha(j)}, & \text{if } j \leq k\\
        0, & \text{otherwise}
    \end{cases}
\end{equation}

Here, $\alpha$ is a permutation of $d$ such that $|[x]_{\alpha(j)}| \geq |[x]_{\alpha(j+1)}|$ for $j\in [1,d-1]$.

\item {$\mathbf{Rand_k}$}: The $k$ percent of gradients are selected based on a random permutation of the list of gradients. $Rand_k: \mathbb{R}^d \times \Omega_k \rightarrow \mathbb{R}^d$ is formulated as below.

\begin{equation}
    [Rand_k(x, \omega)]_j := \begin{cases}
        [x]_j, & \text{if } j \in \omega\\
        0, & \text{otherwise}
    \end{cases}
\end{equation}

Here, $\omega$ is a subset chosen from $\Omega_k$ where, $\Omega_k = \binom{[d]}{k}$ represents set of all the k-element subsets of $[d]$. 
\end{itemize}
In the FL context, clients communicate only a sparse selection of $k$ percent of gradients and their corresponding values to the PS, thereby improving communication efficiency.
\begin{figure}[t]
  \begin{center}
    \includegraphics[scale=0.25]{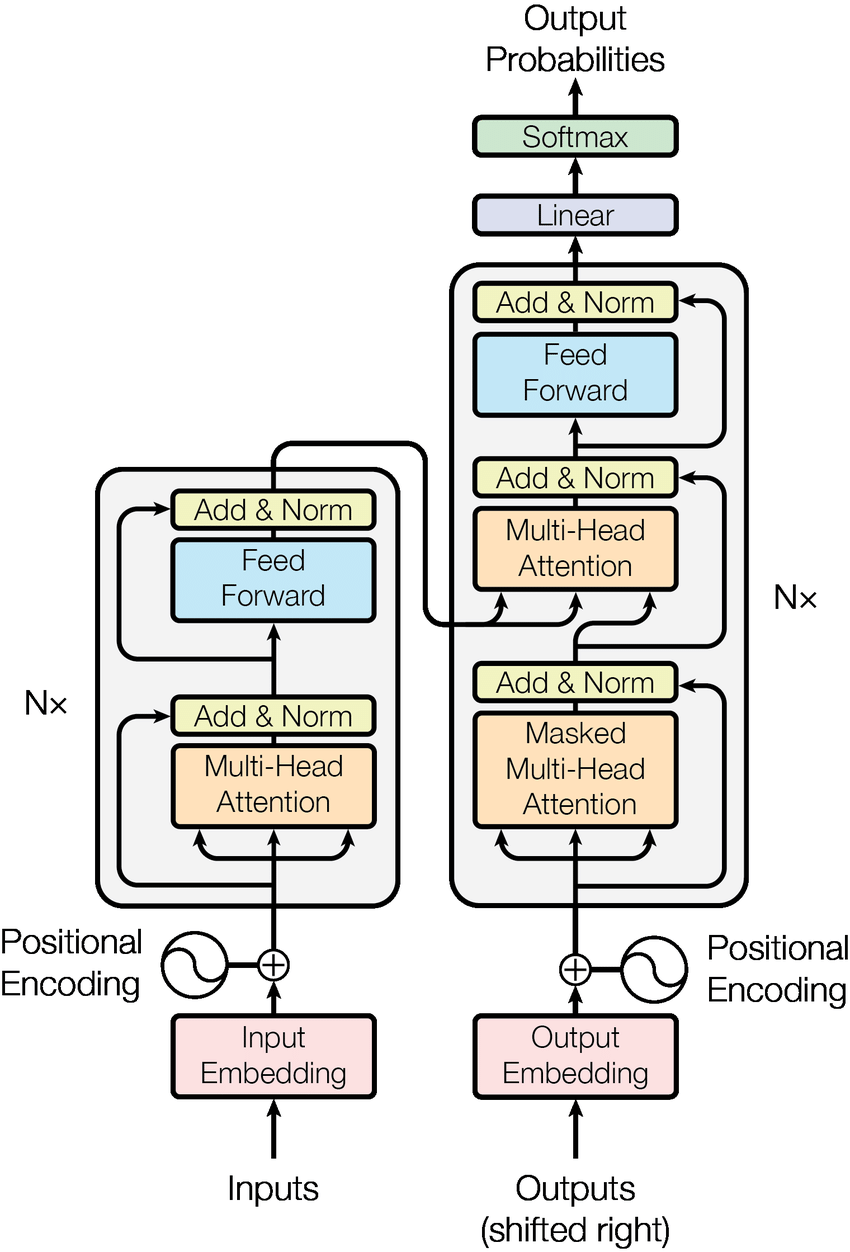}
  \end{center}
  \caption[]{Transformer architecture: encoder on the left block and decoder on the right block.\footnotemark}
  \label{fig:transformer}
\end{figure}
\section{Large Language Models}\label{section3}
LLMs have revolutionized the AI community. They can understand and generate human-like text across various domains. Models like GPT \cite{gpt3, gpt4} have made significant progress in Natural Language Processing (NLP) tasks such as machine translation, question-answering, programming, reasoning, and decision-making. This spectacular performance of LLMs can be attributed to the large number of model parameters trained on a huge text corpus typically extracted from the web. 

The success of LLMs has been further accelerated by large corporations open-sourcing their pre-trained language models for use by the general public. \textit{Llama} \cite{llama} and \textit{Llama2} \cite{llama2} which spans models with 7B to 70B parameters released by \textit{Meta AI}, and \textit{Mistral 7B} \cite{mistral7B}, \textit{Mixtral 8x7B} \cite{mixtral8x7B} by \textit{Mistral.ai} with parameters ranging from 7B to 13B are some of the widely used open-source LLMs in the community.
\subsection{Transformer Architecture}
Every LLM mentioned above occupies some variation of the transformer architecture originally introduced in the paper \cite{vaswani2017attention}. As depicted in Figure \ref{fig:transformer}, the transformer architecture is made up of two blocks: (i) an encoder (shown on the left), and (ii) a decoder (shown on the right).
The transformer architecture is a powerful tool for NLP tasks. Both the encoder and decoder are composed of multiple layers, allowing for the processing and generation of complex language representations. The input text is first tokenized and converted into embedding vectors. Positional encoding is then added to provide information about the position of each token in the sequence, using the following equations
\begin{align}
PE_{(pos, 2i)} &= sin\left(\frac{pos}{10000^{\frac{2i}{d_{model}}}}\right), \\
PE_{(pos, 2i+1)} &= cos\left(\frac{pos}{10000^{\frac{2i}{d_{model}}}}\right).  
\end{align}
Here, $pos$ represents the absolute position of the token in a sequence of tokens, while $i \in [0, d_{model}/2]$ represents the corresponding dimension related to the embedding vector. \footnotetext{The transformer model architecture originally appeared in \cite{vaswani2017attention} with granted permission to reproduce figures for scholarly works.}

The embedded vectors are passed through the multi-head attention layer, which allows the model to focus on different parts of the input simultaneously. The attention mechanism is computed using 
\begin{align}
Attention(Q,K,V) = softmax\left(\frac{QK^T}{\sqrt{d_k}}\right)V,   
\end{align}
where $Q$ is the query vector, $K$ is the key vector with dimensions $d_k$, and $V$ is the value vector with dimension $d_v$ acting as the input to the attention function. 

The output of the attention layer is then passed through a feed-forward layer and residual connections are used to prevent vanishing gradients. The operation of each sub-layer can be summarized as
\begin{align}
output = LayerNorm(x + Function(x)),    
\end{align}
where, $x$ and $output$ represent the input and output of any sub-layer respectively,  while $Function$ can be either Multi-Head Attention or Feed-Forward network function.

The decoder block is similar to the encoder but includes a masked multi-head attention layer to ensure that the model only attends to previous tokens when generating new ones. The output of the decoder is passed through a linear layer and softmax function to predict the probabilities of the next token in the sequence. The transformer architecture can be used in various configurations, including encoder-only, decoder-only, and encoder-decoder models, depending on the specific task at hand.
\begin{figure}[t]
    \centering
    \includegraphics[scale=0.45]{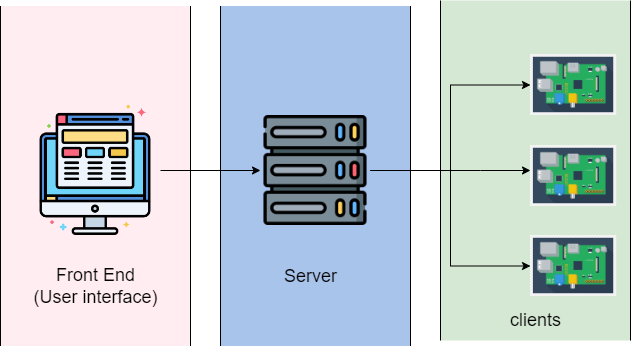}
    \caption{Overview of the web application FL solution.}
    \label{fig:floverview}
\end{figure}
\subsection{Parameter Efficient Fine-tuning of LLMs}
Initially, LLMs are trained on large text corpus which possesses general knowledge. Hence, when adopting an LLM for downstream tasks, fine-tuning the model with a tailored dataset specific to the task at hand may be necessary. Fine-tuning dataset usually comes in the form $\{(x_i, y_i)\}_{i=1}^n $ where $x_i$ is the input to the model and $y_i$ is the target output from the model. Both $x_i$ and $y_i$ are in the form of a sequence of words or tokens. However, fine-tuning an LLM fully with all the parameters included is a huge computationally expensive task requiring large Gigabytes (GB) of Virtual random access memory (VRAM). Therefore, parameter-efficient fine-tuning (PEFT) methods have been explored in the literature.

LoRA \cite{lora} is a widely used method for performing efficient fine-tuning on LLMs allegedly reducing trainable parameters by 10000 folds while lowering the Graphical processing unit (GPU) memory requirement by three times compared to fully fine-tuning a GPT-3 175B parameter model with Adam optimizer. LoRA reduces trainable parameters by decomposing the weight matrices using lower-rank matrices. Consider a pre-trained weight matrix $W \in \mathbb{R}^{d \times k}$. In LoRA, the gradient update is represented as follows
\begin{align}
\Delta W = BA,    
\end{align}
where $A \in \mathbb{R}^{r \times k}$ and $B \in \mathbb{R}^{d \times r}$, with rank $r << \min(d,k)$. The forward pass is calculated as
\begin{align}
h = Wx + \Delta Wx = Wx + BAx.    
\end{align}
where $x$ is the input embedding or intermediate hidden features. Once the fine-tuning is done, inference for the downstream tasks is done by merging the weights as follows
\begin{align}
W = W + BA.    
\end{align}
By this means, adapting LoRA for multiple downstream tasks is possible since it is possible to train different LoRA adapters and plug them into the original model as above.

QLoRA \cite{qlora} is a variant of LoRA to enhance further the efficiency of fine-tuning an LLM. This is achieved by utilizing a quantized version of the transformer model, double quantization, and using paged optimizers. The transformer model is quantized to 4-bit precision using the 4-bit Normal Float (NF4) data type. Hence, QLoRA reduces the GPU memory requirement for fine-tuning a 65B model to less than 48GB, compared to more than 780GB in full fine-tuning.
\section{System Design and Methodology}\label{section4}
As discussed in previous sections, our goal is to implement a web application to make FL accessible for everyone without dealing with the intricacies of network programming to implement a communication architecture. To that end, we develop a new web application with an inbuilt communication architecture to handle FL tasks. By doing so, we eliminate the need for programming an FL solution from scratch. Our solution is mainly divided into three phases
\begin{itemize}
    \item \textbf{Phase 1:} We develop a standard web solution to facilitate FL tasks.
    \item \textbf{Phase 2:} We fine-tune and integrate an LLM to automate the developed solution further.
    \item \textbf{Phase 3:} We explore NAS and HPO using LLM to improve the results of the automated solution further.
\end{itemize}
\begin{algorithm}[t]
\caption{Modified Federated Averaging Algorithm}
\label{alg:fed_avg_new}
\begin{algorithmic}[1]
    \State \textbf{Inputs:} Number of clients $K$, learning rate $\eta$, client fraction $C$, total communication rounds $T$, number of local epochs $E$, mini-batch size $B$. 
    \State \textbf{Output:} Final global model $w_{T+1}$. 
    \State Initialize global model $w_0$
    \For{each round $t = 1, 2, ..., T$}
        \State $m \gets \max(C \cdot K, 1)$
        \State $S_t \gets$ (set of $m$ clients according to the scheduling mechanism selected)
        \For{each client $k \in S_t$ \textbf{in parallel}}
            \State $\hat{w}_{t+1}^k \gets \textbf{ClientUpdate}(k, w_t)$
            \State \textbf{sends} $\hat{w}_{t+1}^k$ to server
        \EndFor
        \If{compress = True}
            \State $w_{t+1}^k \gets \mathbf{DeCompress}(\hat{w}_{t+1}^k)$
        \EndIf
        \State $m_{t} \gets \sum_{k \in S_t} n_k$
        \State $w_{t+1} \gets \sum_{k \in S_t} \frac{n_k}{m_t} w_{t+1}^k$
    \EndFor
\vspace{0.5cm}
    \State \textbf{ClientUpdate}$(k, w)$
    \State $\mathcal{B} \gets$ (split $\mathcal{P}_k$ into batches of size $B$)
    \For{each local epoch $i$ from $1$ to $E$}
        \For{batch $b \in \mathcal{B}$}
            \State $w \gets w - \eta \nabla \ell(w; b)$
        \EndFor
    \EndFor
    \If{compress = True}
        \State $\hat{w}\gets \mathbf{Compress}(w)$
     \EndIf
\end{algorithmic}
\end{algorithm}

\begin{figure}[t]
    \centering
    \includegraphics[scale=0.25]{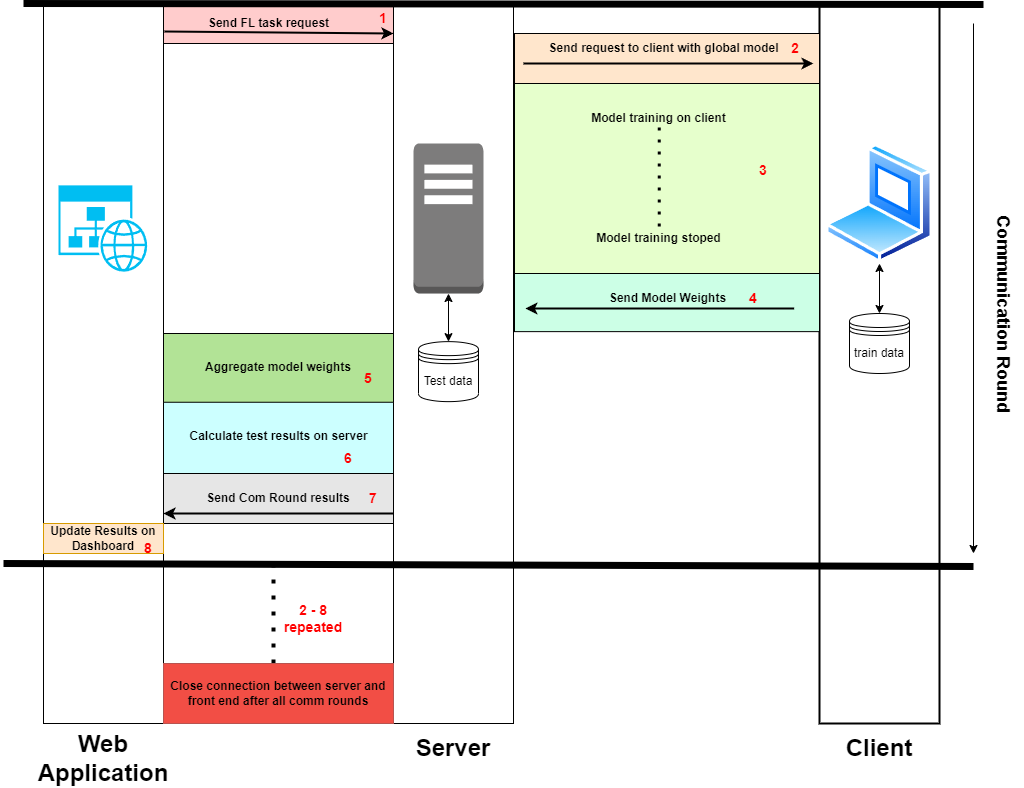}
    \caption{Communication process between Front-end, server, and clients.}
    \label{fig:signaling}
\end{figure}
\subsection{Standard Web Application}
Our proposed solution comprises three key elements, as illustrated in Figure \ref{fig:floverview}. The front-end user interface\footnote{The front-end and back-end codes are publicly available at \url{https://github.com/ICONgroupCWC/FedLFrontEnd} and \url{https://github.com/ICONgroupCWC/FedLBE}, respectively.} is designed to enable end users to input essential parameters required for executing an FL task. The server (here onwards known as FL server) functions as the central coordinator, responsible for tasks such as aggregating parameters, scheduling clients, and managing database services. Additionally, individual clients run separate code locally for training, with each client capable of operating on various hardware setups, provided they possess sufficient computational resources for local training and a communication interface for interacting with the server.

The front-end web user interface is developed as a React \cite{react} application, where Redux \cite{redux} is used for managing the state of the front-end data. We use a web form in the front end to get the necessary information to carry out an FL task. Comprehensive information about the most important fields in the web form to be filled by the end user is explained in Appendix \ref{webfields}. The user has to submit the required task information via the web form, after which the data is converted to Javascript object notation (JSON) format and transmitted to the FL server to initiate the FL task. An example of the content of a JSON can be found in Appendix \ref{JSON}.  

WebSockets serve as the communication protocol for all interactions between the front-end and FL server and between the FL server and the clients. WebSockets is a full duplex communication protocol built on top of the transmission control protocol
(TCP), which facilitates real-time data transfer between two communicating devices. In Appendix \ref{websockets}, we explain how the communication flow of WebSockets and the reasons for adopting this choice of communication protocol.

\subsection{Modified FedAvg}
To accommodate model compression and client scheduling, we make the following changes to the original FedAvg algorithm 
\begin{itemize}
    \item Model compression schemes are implemented for efficient communication of model weights between the client and the FL server. Users can select a compression scheme to compress the locally trained model before transferring the weights back to the PS. The options for the compression scheme are quantization, Top-k or Rand-k.
    \item Several scheduling mechanisms are implemented to select clients for each round of communication. Users can choose from the following options
    \begin{itemize}
        \item \textbf{Random:} PS selects clients randomly based on a given client fraction $C$.
        \item \textbf{Round Robin:} PS selects clients in a round-robin fashion.
        \item \textbf{Latency-proportional:} PS selects clients with the lowest latency averaged over $k$ communication rounds.
        \item \textbf{Full:} PS selects all clients during each communication round.
    \end{itemize}
\end{itemize}
With these additions in mind, we adapt the implementation of the FedAvg as in Algorithm \ref{alg:fed_avg_new} for our web solution. When the value of the variable $compress$ is set to \textit{False} and the scheduling mechanism is \textit{random}, we recover the original FedAvg algorithm. In Algorithm \ref{alg:fed_avg_new}, $\mathbf{Compress}$ and $\mathbf{DeCompress}$ are functions implemented on the client and server sides, respectively, for each compression scheme described earlier. 
\begin{table}[t]
\centering
\caption{Default values to be used for parameters when unspecified in the prompt.}
\begin{tabular}{@{}lc@{}}
\toprule
\textbf{Key} & \textbf{Value} \\ \midrule
minibatch & 16 \\
\midrule
algorithm & classification \\
\midrule
epoch & 5 \\
\midrule
lr & \begin{tabular}[c]{@{}c@{}}0.0001 (Adam)\\ 0.001 (SGD)\\ 0.004 (AdaGrad)\\ 0.0004 (RMSProp)\end{tabular} \\
\midrule
scheduler & full (random if ClientFraction $<$ 1) \\
\midrule
clientFraction & 1 \\
\midrule
minibatchtest & 32 \\
\midrule
comRound & 10 \\
\midrule
optimizer & Adam \\
\midrule
loss & CrossEntropyLoss \\
\midrule
Compress & No \\ \bottomrule
\end{tabular}
  \label{tab:default_values}
\end{table}

As illustrated in Figure \ref{fig:signaling}, the implementation of the modified FedAvg algorithm leverages Web-Sockets for communication, with eight high-level stages. The process begins with the user submitting an FL request through a web form, establishing a WebSocket connection between the front-end web application and the FL server. Then, the server communicates the request to selected clients based on the client fraction and scheduling mechanism, sending the global model and training parameters. Each client trains the model locally and sends the trained weights back to the server asynchronously. The server aggregates the local models to update the global model, calculates test results, and saves the data in a database. Finally, the current communication round's training and test metrics are sent to the front-end via the existing WebSocket connection, updating the web application dashboard in real-time with user-selected plots. This process is repeated iteratively until all requested communication rounds are completed, maintaining the WebSocket connection between the front-end and FL server throughout the training.
\subsection{LLM Finetuning}
The downstream task in our proposed solution is carrying out an FL task based on the user prompt provided. Thus, the LLM needs to be fine-tuned so that the output from the LLM should be compatible with the FL server. In our previous case with the standard web application, the payload in the request from the user contained a JSON, which carries information required for the FL task such as model parameters, optimizer parameters, etc. Hence, it is natural to use the same structure of the JSON as output from the LLM to make the new solution compatible with the previously developed solution.
\begin{figure}[t]
    \centering
    \includegraphics[scale=0.2]{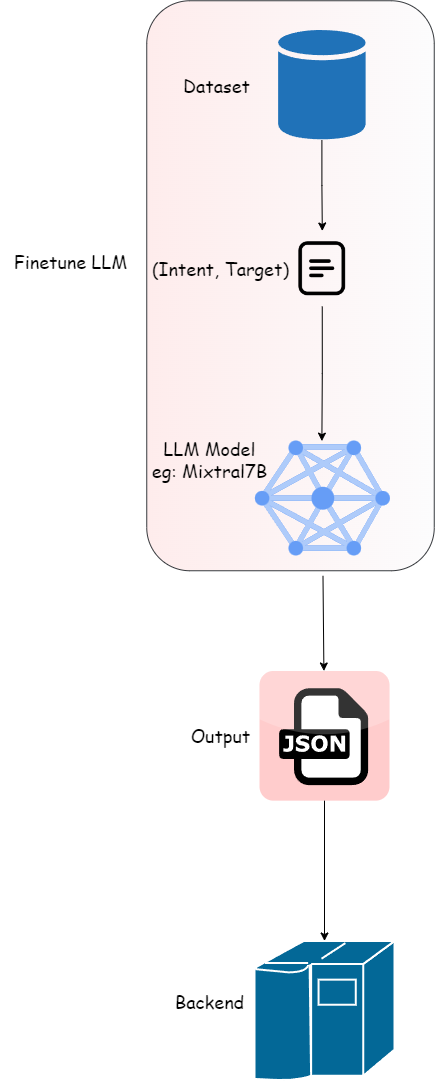}
    \caption{Fine-tuning process of LLM to be compatible with the FL server.}
    \label{fig:llm_finetune}
\end{figure}
As depicted in Figure \ref{fig:llm_finetune}, to fine-tune the LLM, we create a dataset with $<$\textsc{Intent, Target}$>$ pairs, where \textsc{Intent} is the user prompt provided by the user while the \textsc{Target} is the intended JSON output by the LLM for the provided intent. For example, the user intent can be as follows
\begin{quote} 
\textbf{Prompt:} \textit{Create a federated learning task with MNIST dataset. Run for a total of 20 communication rounds and 12 local epochs. Use SGD as the optimizer with a learning rate of 0.004. Use a mini-batch size of 32 for training. The client fraction should be 0.7.} 
\end{quote}
The target JSON should be accustomed according to the provided intent. One caveat here is when the prompt becomes simple, values to be used in the JSON for the unspecified parameters become unclear. To overcome this issue, we use some default values for the parameters whenever the parameter is not specified in the prompt, as specified in Table \ref{tab:default_values}. Some of these default values like learning rate are derived experimentally by running for common datasets used in our work such as MNIST.

One other aspect considered when developing the dataset is possible variation in phrasing the user's intent. For example, when the \texttt{clientFraction} needs to be specified in the intent, it can be phrased in several ways as follows
\begin{quote}
    \textbf{Prompt:}\textit{ Create a federated learning task with MNIST dataset with a client fraction of 0.7.}

    \textbf{Prompt:}\textit{ Perform a federated learning task with MNIST dataset involving 70\% of the total clients.}

    \textbf{Prompt:}\textit{ Execute a federated learning task with MNIST dataset involving 7/10 of the total clients.}

    \textbf{Prompt:}\textit{ Start a federated learning task with MNIST dataset by omitting 30\% of total clients in each communication round.}
\end{quote}
The output JSON of all above prompts should be the same and the dataset is created by including similar variations throughout the dataset for other parameters as well. Apart from that, several other actions are taken to induce variation to the dataset such as paraphrasing, differing complexity of the sentence structure, and differing the order of parameters provided in the intent.

To get good results when fine-tuning an LLM, structuring the prompt is very important. Apart from the user's intent and the target's response, a system instruction is added to the input prompt. The system instruction can provide contextual guidance to the model, informing it about the task at hand or the desired behavior. This helps the model understand the specific requirements of the fine-tuning task and adjust its predictions accordingly. In our case, the System Prompt is used to communicate the role of the LLM succinctly as shown in Appendix \ref{systemprompt}. A detailed description of each parameter of the output JSON is given by the System Prompt for helping to predict the output accurately. The System prompt is kept static throughout all the data points since we are fine-tuning for a single downstream task and the intended output structure is the same for all the user intents. Finally, each data point is customized using the System Prompt, Intent, and Target before feeding into the fine-tuning pipeline.
\subsection{LLM Integration to the Web Application}
\begin{figure}[t]
    \centering
    \includegraphics[scale=0.12]{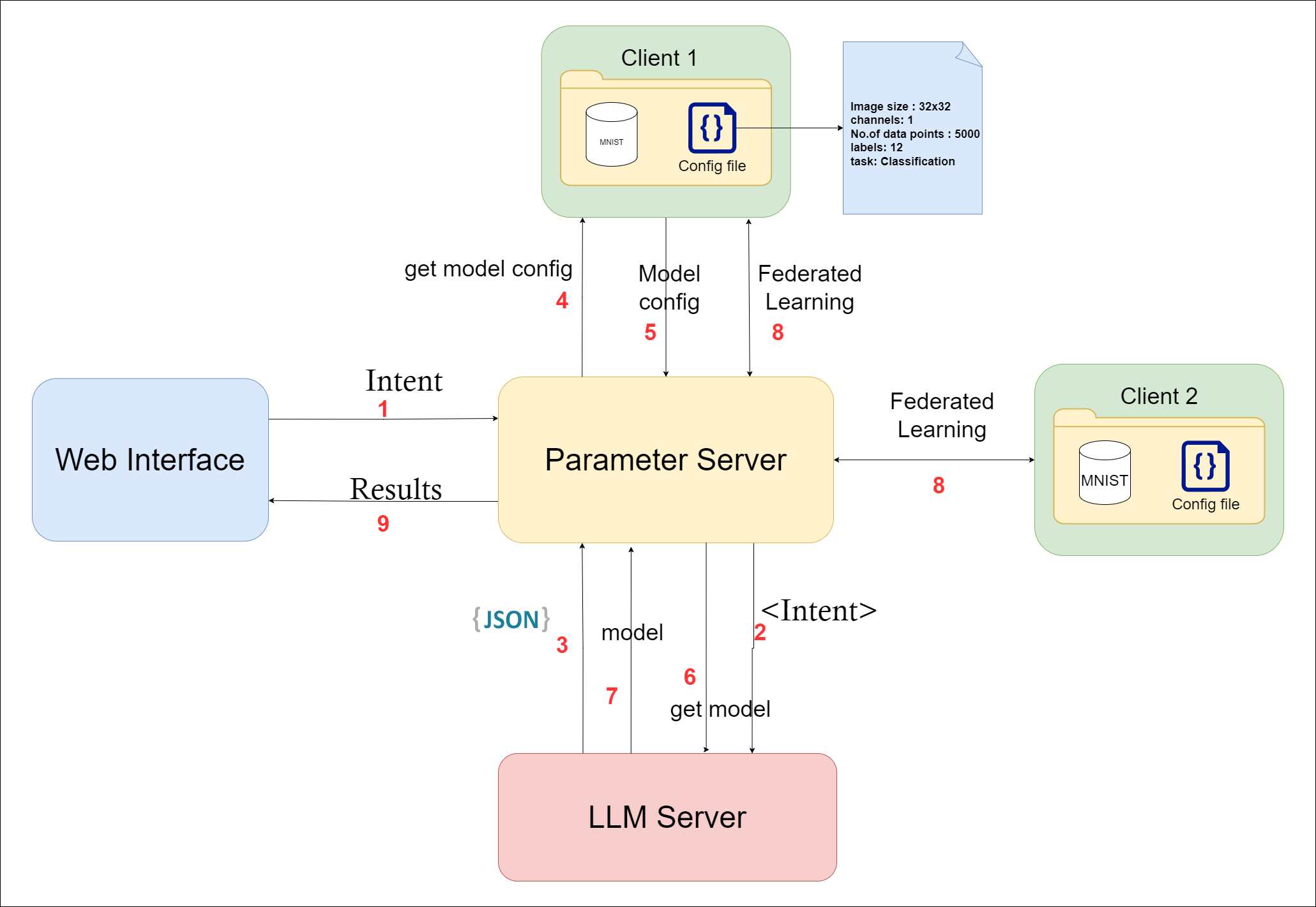}
    \caption{System design for the LLM integrated FL solution.}
    \label{fig:systemdesign}
\end{figure}

Once an open-source LLM model is fine-tuned for our downstream task, the model is integrated into the existing solution. An overview of the extended system design is depicted in Figure \ref{fig:systemdesign}. We implement a separate server\footnote{The code for the LLM server is available at \url{https://github.com/ICONgroupCWC/LLMServer}} to host the LLM model where a Representational State Transfer (REST) API endpoint is created using Python Flask to consume the LLM service. We convert the fine-tuned LLM model to GGUF format before hosting in the LLM Server, where GGUF is a binary file format used for storing LLM models for inference where loading and saving models are made faster. A Python binding \cite{pythonllama} for llama.cpp\cite{llamacpp} is used for serving the model inside the provided REST endpoint.
\begin{figure}[t]
\centering
\begin{subfigure}{.24\textwidth}
  \centering
  \includegraphics[scale= 0.575]{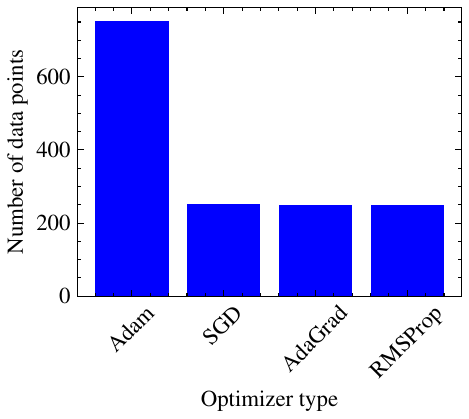}  
  \caption{Training dataset}
\end{subfigure}
\hfill
\begin{subfigure}{.24\textwidth}
  \centering
  \includegraphics[scale=0.55]{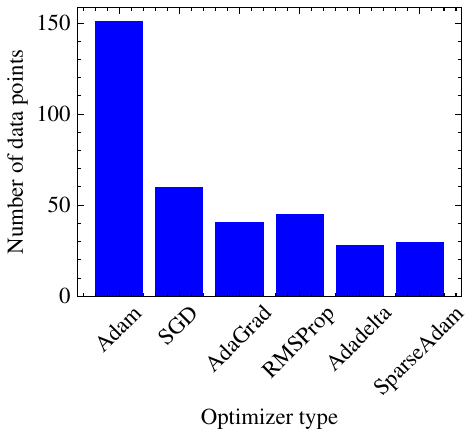} 
  \caption{Test dataset}
\end{subfigure}
\caption{Optimizer variations across the train and test datasets for fine-tuning the LLM.} 
\label{fig:optimizer}
\end{figure}

One challenge for the integrated solution is providing the global model architecture for carrying out the FL task. In the standard web application solution, there is the option of submitting a Pytorch model architecture file. However, given that input for the LLM-integrated solution is in natural language, it is not feasible to provide a model architecture as describing the model architecture in natural language is a complex and tedious task. To overcome this, we utilize the reasoning ability of OpenAI ChatGPT to propose a model architecture for a given task. Through experimental results, we verify that given the input and output dimensions of the network and the number of data points, ChatGPT can provide a model architecture accordingly.
Since data is a commodity of the clients in the FL process, the responsibility of
providing the details of the dataset is given to the clients. A configuration file along
with the dataset should be kept in the data folder, which holds information such as
the dimension, the number of data points, the task, and the number of labels as depicted in Figure \ref{fig:systemdesign}. Using this provided configuration file a new prompt is created to query ChatGPT for a model architecture as follows.
\begin{quote}
\textbf{User\_Prompt:}\textit{ Create a model architecture for the following task. The task is a \underline{Classification} task with \underline{12} labels. The input data is a tensor of shape \underline{(1,1,32,32)} and has \underline{5000} datapoints altogether. Considering the above information, create a neural network architecture that could achieve good accuracy.}
\end{quote}
The above user prompt is always kept static, with placeholders for underlined values where the underlined values are overridden using the configuration file provided in the client data folder. Similar to the previous case, a system prompt with task details is used to support this prompt before the request is sent to the ChatGPT API. Accordingly, we implement a separate REST endpoint in the LLM server to query a model architecture for a given task.

As illustrated in Figure \ref{fig:systemdesign}, the communication process of the integrated solution occurs in nine high-level stages. Initially, the user sends the intent for the FL task to the PS. Next, the PS forwards the intent to the LLM server endpoint to get the configuration of the FL task, where the LLM server predicts the configuration using the fine-tuned LLM and returns the configuration file to the PS as the next step. Then, the PS requests a randomly selected client for the configuration file which contains the dataset's properties. Next, the returned configuration file is forwarded to the LLM server, where the LLM server overrides the aforementioned user prompt and queries the ChatGPT API to provide a model architecture suitable for the task. Once the LLM server receives the model architecture, it is returned to the PS by the LLM server. Finally, the FL process is carried out as usual using the received model architecture as the global model and intermediate results are communicated with the front-end.   
\subsection{Neural Architecture Search and Hyperparameter Optimization}
\begin{figure}[t]
    \centering
    \includegraphics[scale=0.11]{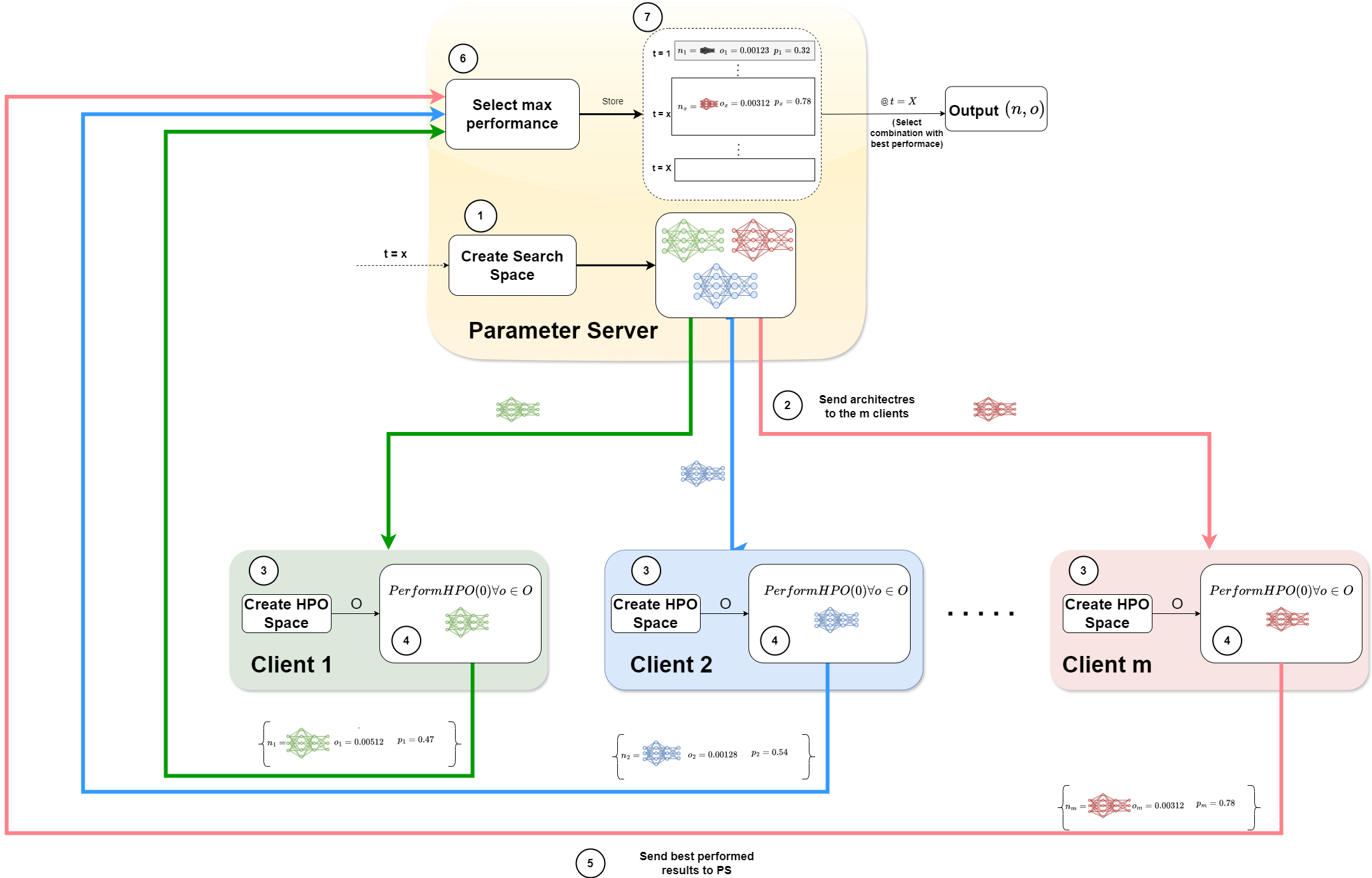}
    \caption{Federated NAS and HPO process.}
    \label{fig:nas_hpo}
\end{figure}

As discussed in the previous section, we observe that OpenAI ChatGPT has the ability to suggest a neural network architecture, given the nature of the task and the dataset. However, one caveat in the proposed methodology is achieving good test accuracy is uncertain since the task is carried out using an arbitrary model suggested by the ChatGPT. Hence, in this section, we focus on improving the test accuracy by leveraging ChatGPT to generate the search space for NAS. Also, we perform HPO on the suggested models in the search space to find the best hyperparameters suited for the respective models. Both NAS and HPO are performed in a distributed manner before the FL task starts. The procedure we follow in performing NAS and HPO is provided in Algorithm \ref{alg:nas_hpo} and illustrated in Figure \ref{fig:nas_hpo}.

According to the provided algorithm, NAS is performed for $X$ number of search rounds as provided in the input. We utilize ChatGPT to initialize the search space at each round, where the \textbf{CreateSearchSpace} function will produce $m$ number of models where $m$ is the total number of clients participating in the training at the current search round. Two prompts are used in generating model architectures where at the first round \textbf{InitialPrompt} is used while at later rounds \textbf{IntermediatePrompt} is used. Also, an \textbf{ErrorPrompt} is used to handle errors whenever the models produced by the ChatGPT are invalid. The prompts used are as follows: For all prompts, the underlined values are given as input to the function. 

\begin{quote} 
\textbf{InitialPrompt:} \textit{You are an AI that strictly conforms to responses in Python. You are an assistant in providing CNN architectures as a search space for neural architecture search. I need help designing the optimal CNN architecture for a specific dataset. I plan to start with a variety of models and change the model architecture based on model performance. As the initial step, please provide \underline{3} CNN architectures with varying width, depth, and layer types. The input data is in the form \underline{torch.zeros((1,3,32,32))} and the output is in the form \underline{torch.size((1,10))}. There are altogether \underline{10000} data points for training. When suggesting models give importance to the above factors to make the model complex accordingly. After you suggest a design, I will test its performance and provide feedback. Based on the results of previous experiments, we can collaborate to iterate and improve the design. Please avoid suggesting the same design again during this iterative process. Your responses should contain valid Python code only, with no additional comments, explanations, or dialogue. Provide the PyTorch models according to the given prompt. Important!! Output the solution as \underline{3} different Python code bases. Start each code base with $<$Code$>$ and end the code base with $<$$/$Code$>$ brackets. So there must be \underline{3} brackets. Include PyTorch imports with `import torch', `from torch import nn', and `import torch.nn.functional as F'. Do not include model initialization code.} 
\end{quote}

\begin{quote}
    \textbf{Intermediate Prompt: } \textit{By using provided model {model}, we achieved an accuracy of \underline{57}\%. As previously Please recommend \underline{3} new models that outperform prior architectures based on the above-mentioned experiments on neural architecture search. Take a step towards making the model more complex by adding more layers to the neural network and increasing the number of parameters in the model.  The output should be structured the same as in the previous case. As in the previous case, the input is in the form \underline{torch.zeros((1,3,32,32))}, and the output is in the form \underline{torch.size((1,10))}. Please make sure that all the dimensions are correct in the network and no error will arise}
\end{quote}

\begin{quote}
    \textbf{Error Prompt: } \textit{The suggested model \underline{(model)} gives the following error \underline{(error message)}. Please suggest a new model architecture with similar parameter complexity that conforms with the dimensions of the input and output correctly to avoid the above error. The input is in the form \underline{torch.zeros((1,1,32,32))} and output a tensor in the form \underline{torch.size((1,10))}}
\end{quote}

During each search round, the produced $m$ model architectures are sent to the $m$ number of local clients to carry out HPO. For each client, HPO is run for $H$ number of rounds, where at each round, $Y$ hyperparameter configurations are created. In our study, we only focus on learning rate as a hyperparameter as we believe learning rate has the most effect on model performance. Furthermore, focusing on a single hyperparameter significantly reduces the search space. This simplification allows for a more thorough exploration of learning rate values within the given computational constraints. Hence, $Y$ number of learning rates between a provided upper bound of $5 \times 10^{-2}$ and a lower bound of $10^{-5}$ will be generated randomly by \textbf{CreateHPOSpace} function at each HPO round by each local client.

\begin{algorithm}[H]
\caption{Algorithm for NAS and HPO for automated solution}
\label{alg:nas_hpo}
\begin{algorithmic}[1]
    \State \textbf{Inputs:} Number of search rounds $X$, number of HPO rounds $H$, number of explorer epochs $E$, total number of clients $K$, client fraction $C$, Hyperparameter types $T$, number of hyperparameters $Y$, data configurations $d$, initial batch size $B$ 
    \State \textbf{Output:} Final best model $n$, Best hyperparameters $o$ 
    
    \For{each round $t = 1, 2, ..., X$}
        
        \State $m \gets \max(C \cdot K, 1)$
        \State $N_t \gets \textbf{CreateSearchSpace}(m, d)$
        \State $S_t \gets$ (set of $m$ clients selected randomly)
        \For{each client $k \in S_t$ \textbf{in parallel}}
            \State $n_k \in N_t \gets$ (receives model architecture from Server)
            \For{each round $j = 1, 2, ..., H$}
                \State $O \gets \textbf{CreateHPOSpace}(T, Y)$
                \State $o_j, p_j \gets \textbf{PerformHPO}(n_k, O, E, B)$
              
            \EndFor
            
            \State $o_k, p_k \gets$ (where $ k = \arg \max p_j$)
            \State \textbf{sends} $(n_k, o_k, p_k)$ to server
        \EndFor
        \State \{$(n_k, o_k, p_k) \mid k = 1, 2, 3, \ldots, m $\} (receives from clients)
        \State $(n_t,o_t, p_t) \gets ( \text{where }  t =  \arg \underset{1 \leq k \leq m} \max p_k$ ) 
    
    \EndFor
    \State $(n_f,o_f, p_f) \gets ( \text{where }  f =  \arg \underset{1 \leq t \leq X} \max p_t$ )
    \State $n,o \gets (n_f, o_f)  $

    \vspace{0.5cm}
    \State \textbf{PerformHPO}$(n, O, E, B)$
    \State $p_b = 0$ // Stores best performance
    \State $o_b = \{\} $ // Stores best hyperparameter combination 
    
    \State $current\_round = 0$
    \While{$O.length > 1$}
    \If{$current\_round \neq 0$ }
        \State $O \gets$ (Select the top performing half of hyperparameters)
        \State $B \gets min(B \cdot 2, traindata.size)$
    \EndIf
    \State $\beta \gets $ (select $B$ datapoints randomly from training data)
        \For{each combination $o \in O$}
            \For{each epoch $e = 1, 2, 3 \ldots , E $}
                \State \textbf{TrainModel}$(n,o)$
            \EndFor
            \State $p \gets \textbf{EvaluateModel}(n,o, \beta)$
            \If{$p > p_b$}
                \State $p_b \gets p$
                \State $o_b \gets o$
            \EndIf
            
        \EndFor
        \State $current\_round += 1$
    \EndWhile
    \State return $(o_b, p_b)$
    
\end{algorithmic}
\end{algorithm}
\newpage
The created HPO space is passed to the \textbf{PerformHPO} function where the HPO operation is carried out using a selective halving manner. At each step after the first step, hyperparameter space is reduced by half size where only the best $50\%$ of the hyperparameters per their performance are retained. Meanwhile, the training set size is also increased by two folds to ensure the retained hyperparameter space is trained on more data points while the HPO progresses. The model architecture received by the local client is trained for $E$ number of epochs, for each retained hyperparameter using the batch of data $\beta$ in the local client.

Once the training is done, the model is evaluated using a validation dataset to find the best-performing hyperparameter. Here, we use the validation accuracy as the performance metric. The best candidate is updated at each round where the \textbf{PerformHPO} process is carried out until only one candidate from the HPO space is retained with the highest performance. This procedure is run for $H$ number of HPO operations where the hyperparameter and performance pairs will be recorded. At the end of all the $H$ HPO rounds, the best-performing hyperparameter will be selected from the pool of recorded hyperparameters. Then, the combination of the model, best hyperparameter, and its performance is sent to the server where the server stores these combinations in a list for future comparison. Once all the NAS rounds are completed, the model and hyperparameter combination with the best performance is selected as $(n,o)$ to be used as the global model and learning rate for the specific FL process.

\begin{figure}[t]
\centering
\includegraphics[scale= 0.25]{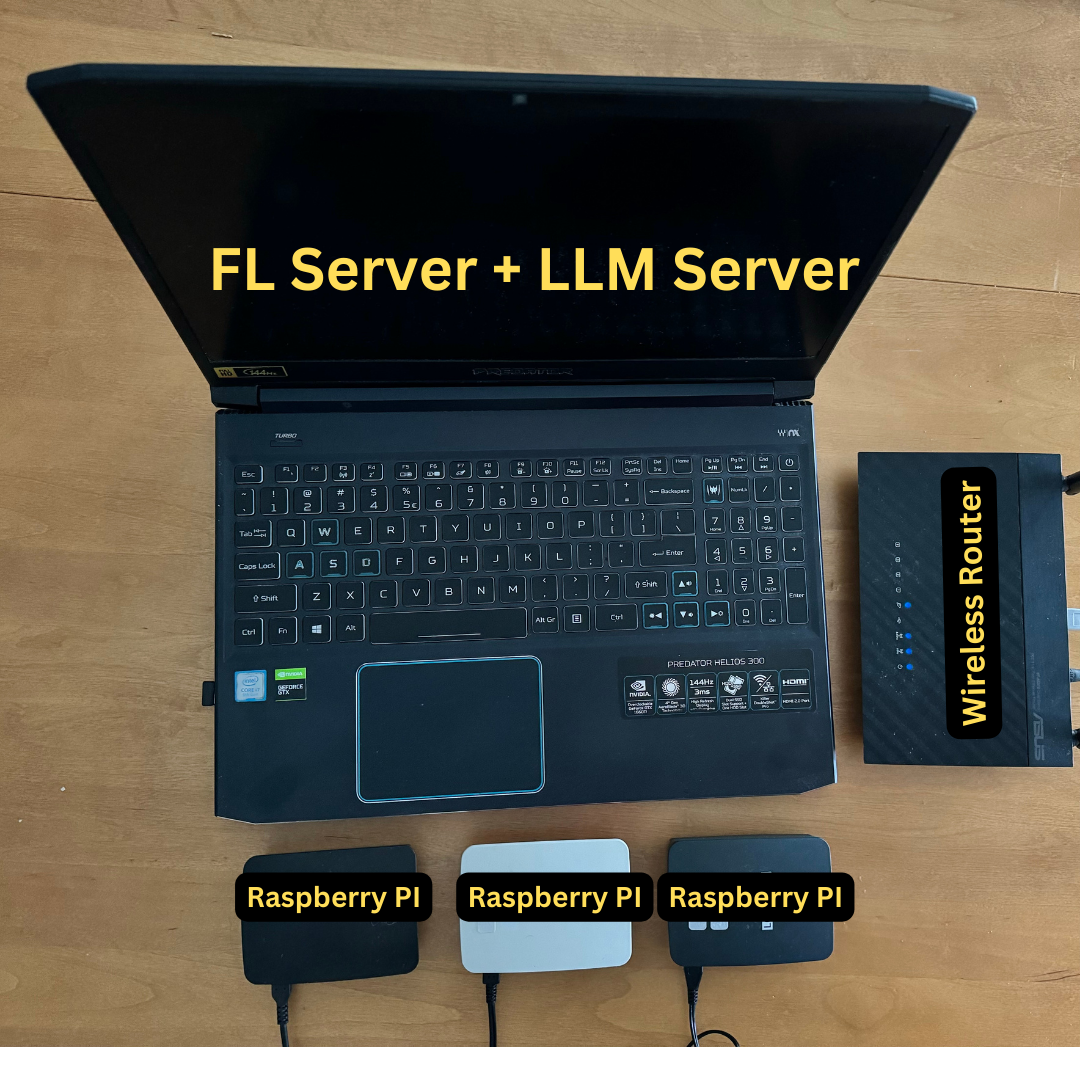}  
\caption{Experimental setup: three Raspberry PIs, a laptop, and a wireless router.} 
\label{fig:hardware}
\end{figure}
\begin{table}[t]
\centering
\caption{A summary of the different use cases used in our experiments using MNIST, Fashion-MNIST, CIFAR-10, and SVHN datasets.}
\begin{tabular}{@{}lp{7cm}@{}}
\toprule
\textbf{Cases} & \textbf{Description} \\ \midrule
Case 1         &  An FL task using \textbf{MNIST} dataset, with a client fraction = 0.7, local minibatch size = 4, number of local epochs = 1, communication rounds = 50, optimizer = Adam, learning rate = 0.0001                  \\
\midrule
Case 2         &  An FL task using \textbf{CIFAR-10} dataset with a client fraction = 0.7, local minibatch size = 8, number of local epochs = 2, communication rounds = 100, and optimizer = Adam, learning rate = 0.001.                  \\
\midrule
Case 3         &  An FL task using \textbf{Fashion-MNIST} dataset, with a client fraction = 0.8, local minibatch size = 8, number of local epochs = 1, communication rounds = 50, optimizer = Adam, learning rate = 0.0001                    \\
\midrule
Case 4         &  An FL task using \textbf{SVHN} dataset with a client fraction = 0.7, local minibatch size = 8, number of local epochs = 1, communication rounds = 50, and optimizer = Adam, learning rate = 0.0001.                    \\
\bottomrule
\end{tabular}
\label{tab:cases}
\end{table}
\section{Experimental Results}\label{sec:numerical}\label{section5}
\subsection{LLM Finetuning Results}
For the fine-tuning part, we start by creating a new dataset with 1504 data points with $<$\textsc{Intent, Target}$>$ pairs for fine-tuning the LLM\footnote{The fine-tuned dataset is available at \url{https://huggingface.co/datasets/cmarvolo/auto_fl}.}. The dataset is created while considering a wider range of possible variations that could occur in the user intent. As illustrated in Figure \ref{fig:optimizer}, several variations of optimizer choices are included in the dataset. As observed in the plots, the number of instances containing default values is higher. This is due to the higher number of data points not specifying the relevant parameter in the intent. For example, it is natural for a user to input only the \texttt{clientFraction}, but not the \texttt{scheduler}, where in such cases the default value is used.

This dataset is used to fine-tune a Mixtral-7B\cite{mistral7B} model. The fine-tuning process is carried out for two epochs through the dataset using the QLoRA method\footnote{The fine-tuned LoRA adapter is available at \url{https://huggingface.co/cmarvolo/mistral-7b-fed-auto-lora}.}. Hence, the model is loaded in 4-bit precision for fine-tuning, where at the end of the training, the LoRA module is merged with the base model by upscaling the trained LoRA to 8-bit precision. We use an 8-bit precision variation of adamW \cite{adamw} optimizer, Adamw\_8bit, as the optimizer with a learning rate $2 \times 10^{-4}$ for fine-tuning. A single A100 GPU provided by Google Colab is used for fine-tuning where Unsloth \cite{Unsloth} library is used for efficient fine-tuning. Unsloth is an open-source library that provides Python API for faster fine-tuning of LLMs.


To evaluate the performance of the fine-tuned Language Model, we create a new test dataset comprising 355 instances similar to the fine-tuned dataset. The dataset is designed to span from simpler prompts with minimal specified parameters to complex prompts with many parameters included, ensuring a comprehensive evaluation spectrum. Additionally, for categorical parameters such as optimizers, new values were added to test the generalization capability of the LLM as evident from Figure \ref{fig:optimizer}, where Adadelta and SparseAdam were not used as optimizers in the intents in the fine-tuning stage. Similarly, values for the \textit{dataset} key were also given unseen values with random dataset names to evaluate the generalizability.

The evaluation accuracy of the test dataset holds at $99.4\%$, where only two data points are incorrectly predicted. 

\begin{figure}[t]
\centering
\includegraphics[scale= 0.65]{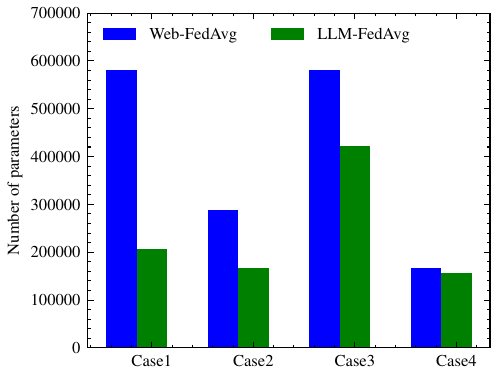}  
\caption{Number of parameters of the model used for Web-FedAvg and LLM-FedAvg for each case.} 
\label{fig:modelparameters}
\end{figure}
\subsection{Comparative Analysis of Web-FedAvg and LLM-FedAvg Implementations}
Several simulations are carried out using both standard and LLM-integrated solutions to evaluate the results. In the remainder of this paper, we denote by \textit{Web-FedAvg} and \textit{LLM-FedAvg} the standard solution and the LLM-integrated solution, respectively. For all the simulations, three Raspberry Pi modules are used as clients and both the PS and the LLM server are run on a single Laptop with an i7-9750H CPU, 24GB RAM, and a NVIDIA Geoforce 1660Ti GPU with 6GB dedicated memory, where all the devices are connected to a single local wireless network. The hardware setup used in the simulations is illustrated in Figure \ref{fig:hardware}. 

Simulations are run for several cases using MNIST, CIFAR-10, Fashion-MNIST, and SVHN datasets equally distributed among the clients. A summary of scenarios used for simulations is given in Table \ref{tab:cases}. For the cases represented in the table, \textit{LLM-FedAvg} method is carried out by creating a relevant prompt to capture the necessary parameters. The simulation results are tabulated in Table \ref{tab:results} for both \textit{Web-FedAvg} and \textit{LLM-FedAvg} methods, where communication overhead and CPU time represent the total number of bytes transferred from the client to the server and the total elapsed time for all the communication rounds, respectively. The number of parameters of the models used by both methods is plotted in Figure \ref{fig:modelparameters}, where for \textit{Web-FedAvg}, a handcrafted model is used, while for \textit{LLM-FedAvg}, the model is suggested by ChatGPT given the dataset configuration.

From case 1, it is observed that both cases achieve test accuracy over 98\% where \textit{Web-FedAvg} achieves slightly better accuracy. However, communication overhead and CPU time are much less for \textit{LLM-FedAvg}. For case 2, \textit{LLM-FedAvg} seems to achieve around 3\% better test accuracy than \textit{Web-FedAvg} while maintaining a lesser communication overhead and CPU time. Observing case 3, we can see that both \textit{Web-FedAvg} and \textit{LLM-FedAvg} achieve a similar accuracy of around 89\% with \textit{LLM-FedAvg} achieving the results with lesser communication overhead and CPU time. Finally, in case 4, \textit{LLM-FedAvg} manages to achieve around 3\% better accuracy over \textit{Web-FedAvg} while having lesser communication overhead and elapsed CPU time. In summary, both \textit{Web-FedAvg} and \textit{LLM-FedAvg} seem to achieve similar accuracy for simulations carried out, while sometimes \textit{LLM-FedAvg} achieve slightly better accuracy. However, for all the cases communication overhead and elapsed CPU time for the FL process is lesser in \textit{LLM-FedAvg} than \textit{Web-FedAvg}. The reason is that chatGPT continuously suggests models with lesser parameters as depicted in Figure \ref{fig:modelparameters} for \textit{LLM-FedAvg} for all the cases.

\begin{table}[t]
\caption{Performance of Web-FedAvg and LLM-FedAvg for the different cases.}
\begin{tabular}{@{}lcccc@{}}
\toprule
\textbf{Cases}          & \textbf{Method} & \textbf{\begin{tabular}[c]{@{}c@{}}Test \\ Accuracy\end{tabular}} & \textbf{\begin{tabular}[c]{@{}c@{}}Communication \\ Overhead (bytes)\end{tabular}} & \textbf{CPU (s)} \\ \midrule
\multirow{2}{*}{Case 1} & Web-FedAvg      &   98.69\%                     &     1.165 $\times 10^2$                                   &   $ 3.761 \times 10^3$               \\
                        & LLM-FedAvg      &   98.14\%                     &     4.150 $\times$ 10                                      &   $2.004 \times 10^3$          \\
\midrule
\multirow{2}{*}{Case 2} & Web-FedAvg      &   69.18\%                    &       6.700 $\times$ 10                                      &   $4.290 \times 10^4$               \\
                        & LLM-FedAvg      &   72.55\%                    &       6.276 $\times$ 10                                    &   $3.880 \times 10^4$               \\
\midrule
\multirow{2}{*}{Case 3} & Web-FedAvg      &   89.74\%                     &            1.165 $\times 10^2$                            &   1.736 $\times 10^3$               \\
                        & LLM-FedAvg      &   89.69\%                     &             8.450 $\times 10$                            &  1.649 $\times 10^3$                \\
\midrule
\multirow{2}{*}{Case 4} & Web-FedAvg      &   86.30\%                     &                    3.350 $\times 10$                    &   1.643 $\times 10^4$               \\
                        & LLM-FedAvg      &   89.87\%                     &      3.150 $\times 10$                                  &  1.583 $\times 10^4$                 \\
\bottomrule
\end{tabular}
\label{tab:results}
\end{table}

\subsection{Comparative Analysis of LLM-FedAvg and LLM-FedAvgNAS Implementations}
Although \textit{LLM-FedAvg} manages to achieve a considerably good test accuracy, we believe the accuracy could be further improved. Hence, we utilize NAS and HPO as described in Algorithm \ref{alg:nas_hpo}, termed as \textit{LLM-FedAvgNAS} here onwards. We only perform NAS and HPO on CIFAR-10, Fashion-MNIST, and SVHN datasets where MNIST is omitted as a higher accuracy over 98\% is already achieved.  For all the \textit{LLM-FedAvgNAS} search processes, five search rounds ($S$) with two local HPO rounds ($H$) are carried out. Only the learning rate is used as the hyperparameter type ($T$). The total number of epochs ($E$) is kept as 20 while a client fraction of one is set to use all three clients in the search process. The number of hyperparameters ($Z$) is set as 20, where for each HPO round, 20 random learning rates are created as HPO search space. The initial batch size for all the simulations is kept at 250. Due to the increased compute and memory demand for the search process, the clients, in this case, are run using a single V100 GPU since Raspberry PI compute is insufficient to carry out the search process.

\begin{figure}[t]
\centering
\begin{subfigure}{.24\textwidth}
    \centering
    \includegraphics[scale= 0.6]{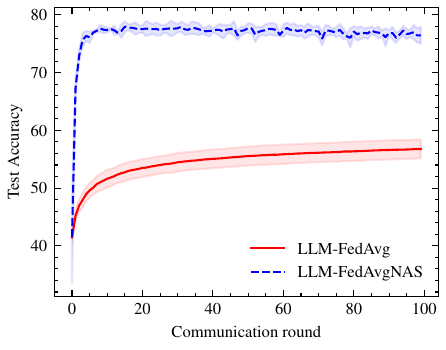} 
    \caption{CIFAR-10 dataset}
\end{subfigure}
 \hfill
 \begin{subfigure}{.24\textwidth}
    \centering
    \includegraphics[scale= 0.6]{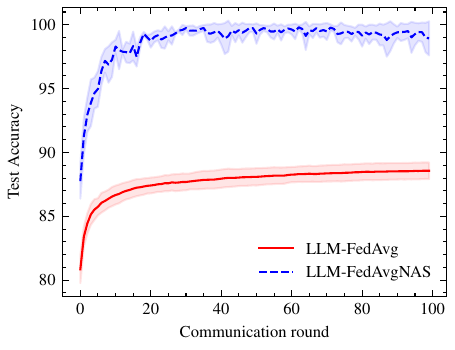} 
    \caption{Fashion-MNIST dataset}
 \end{subfigure}
 \hfill
 \begin{subfigure}{.24\textwidth}
    \centering
    \includegraphics[scale= 0.6]{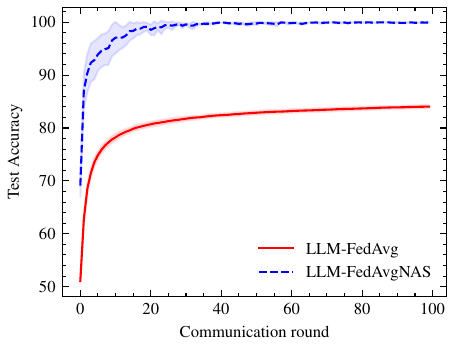} 
    \caption{SVHN dataset}
 \end{subfigure}
 
\caption{Results comparison between LLM-FedAvg and LLM-FedAvgNAS for CIFAR-10, Fashion-MNIST and SVHN datasets.} 
\label{fig:cfar10nas}
\end{figure}

Once the best model and learning rate are obtained using \textit{LLM-FedAvgNAS} search, the FL process is carried out separately using the obtained results for 100 communication rounds and only one local epoch with a client fraction of 0.7 to utilize only two of three total clients. \textit{LLM-FedAvgNAS} search is carried out three times obtaining three global model and learning rate pairs to plot the mean and standard deviation of the test accuracy of FL communication rounds as illustrated in Figure \ref{fig:cfar10nas}. Similarly, \textit{LLM-FedAvg} is carried out three times using the same number of communication rounds and client fraction. The prompt used for this only specifies the dataset, communication rounds, local epochs, and client fraction ensuring that default values specified in Table \ref{tab:default_values} are used for the rest of the parameters. Similar to the previous case, the mean test accuracy and standard deviation for \textit{LLM-FedAvg} simulations are plotted in Figure \ref{fig:cfar10nas}.

Observing the results in Figure \ref{fig:cfar10nas}, we can see that \textit{LLM-FedAvgNAS} surpasses in test accuracy by a large margin \textit{LLM-FedAvg}. For the CIFAR-10 dataset, \textit{LLM-FedAvgNAS} manages to achieve a test accuracy of around 76\% while for Fashion-MNIST and SVHN datasets, an optimal accuracy of over 98\% is achieved. Hence, we conclude that although the \textit{LLM-FedAvg} method provides decent results, incorporating NAS using LLM and performing HPO, the performance of the provided automated solution can be improved further. Although \textit{LLM-FedAvgNAS} offers better performance, it comes with a significant computational overhead during the offline search process before starting the FL task as illustrated in Table \ref{tab:nastimes}, which summarizes the average CPU time taken for a single search round for each dataset.

\begin{table}[t]
\caption{Average times for single search round in \textit{LLM-FedAvgNAS} for different datasets.}
\centering
\begin{tabular}{@{}lc@{}}
\toprule
\textbf{Dataset}  & \textbf{\begin{tabular}[c]{@{}c@{}}CPU(s)\end{tabular}}  \\ \midrule
CIFAR-10 & 1.7651 $\times 10^3$ \\ 
Fashion-MNIST & $1.2127 \times 10^3$ \\
SVHN & $1.7995 \times 10^3$ \\
\bottomrule
\end{tabular}
\label{tab:nastimes}
\end{table}

\section{Conclusion}\label{section6}
In this work, we introduced a web-based solution for orchestrating FL tasks, designed to be accessible to researchers without extensive programming knowledge. Complex communications for seamless support for FL tasks are implemented using an efficient WebSocket architecture. Moreover, the integration of intent-based automation, achieved through fine-tuning an LLM, further enhances user accessibility by allowing FL processes to be automated via natural language prompts. Experimental results demonstrate that our automated solution achieves comparable or superior performance to standard methods while maintaining the communication overhead up to 64\% lower and CPU time up to 46\% lower for the considered scenarios.  We also integrated NAS and HPO to enhance performance.  Notably, the incorporation of NAS and HPO led to significant performance improvements where we observed a 10-20\% test accuracy increase over the test cases run, effectively addressing potential limitations of the automated approach, although with an added computational overhead in the search process.


\bibliographystyle{IEEEtran}
\bibliography{references}
\section{Appendices}
\subsection{Web Form Fields}\label{webfields}
The input parameters by the end user to the web form and their description are shown in the following table.
\begin{table}[H]
\centering
\caption{Fields used in the web form and their description.}
\begin{tabular}{@{}lc@{}}
\toprule
\textbf{Field Name} & \textbf{Description} \\ \midrule
Task Name (\textit{text}) & Unique identifier for the task \\
\midrule
Host IP (\textit{text}) & IP address of the PS \\
\midrule
Client list (\textit{text}) & List of client IP addresses \\
\midrule
Algorithm type (\textit{dropdown}) & \begin{tabular}[c]{@{}c@{}}Classification\\Regression\\ \end{tabular} \\
\midrule
Local minibatch size (\textit{int}) & Minibatch size for local training \\
\midrule
Local epochs number (\textit{int}) & Number of local training iterations \\
\midrule
Learning rate (\textit{decimal}) & Learning rate used for training \\
\midrule
Scheduler Type (\textit{dropdown}) & \begin{tabular}[c]{@{}c@{}}Random\\Round robin\\Latency proportional\\Full \end{tabular} \\
\midrule
Client Fraction (\textit{decimal}) & \begin{tabular}[c]{@{}c@{}}Fraction of clients participating\\in a single round of communication \end{tabular} \\
\midrule
Test Minibatch size (\textit{int}) & Minibatch size for testing \\
\midrule
Communication rounds number (\textit{int}) & \begin{tabular}[c]
{@{}c@{}}Total number of rounds between\\ the PS and the clients\end{tabular}\\
\midrule
Model file (\textit{file}) & Python model architecture file \\
\midrule
Optimizer (\textit{dropdown}) & \begin{tabular}[c]{@{}c@{}}Adam\\SGD\\\dots\\ \end{tabular} \\
\midrule
Loss function (\textit{dropdown}) & \begin{tabular}[c]{@{}c@{}}Cross entropy loss\\MSE loss\\\dots\\ \end{tabular} \\
\midrule
Compression scheme (\textit{dropdown}) & \begin{tabular}[c]{@{}c@{}}Quantize\\Topk\\Randomk\\No compression\\ \end{tabular} 
\\ \bottomrule
\end{tabular}
  \label{tab:webform}
\end{table}
\subsection{JSON File Content}\label{JSON}
The LLM should produce a compatible JSON output when a user needs to run an FL task through an intent-based method. The associated JSON file for this task is given as follows 
\begin{quote}
\begin{lstlisting}[language=json,firstnumber=1]
{"algo": "Classification", 
 "minibatch": "16", 
 "epoch": "5", 
 "lr": "0.0001", 
 "scheduler": "full", 
 "clientFraction": "1", 
 "minibatchtest": "32", 
 "comRounds": "10", 
 "optimizer": "Adam", 
 "loss": "CrossEntropyLoss", 
 "compress": "No", 
 "dataset": "MNIST"}
\end{lstlisting}
\end{quote}
\subsection{Communication Flow of WebSockets}\label{websockets}
Unlike in HTTP protocol, WebSocket connection is long-lived, where the connection is closed only when one party deliberately closes the connection or the connection times out due to idle data flow. Initially, an HTTP request is sent to the server from the client requesting a connection upgrade to WebSockets. Websockets connection is established upon connection upgrade response from the server where the connection is kept until the connection is closed by either party.
The choice of adopting Websockets for our use case is primarily due to,
\begin{itemize}
    \item \textbf{Bi-Directional data transfer:} WebSockets enable bi-directional communication between the Server and the Client. Since in an FL setting, model weights must be communicated between the Server and Client and vice versa, this bi-directional communication enables efficient synchronization of model updates. 
    \item \textbf{Real-time communication:} WebSockets facilitate real-time communication by maintaining a persistent connection between the client and the server. Unlike traditional  Hypertext Transfer Protocol (HTTP), which requires establishing a new connection for each request, WebSockets leverage an established connection, eliminating the overhead associated with connection setup. This real-time communication is advantageous for FL, where rapid exchange of model updates is crucial for collaborative learning.
    \item \textbf{Long-lived connection:} WebSockets connection is long-lived, meaning the connection will exist until a client deliberately closes the connection or the connection is inactive for a longer period. This allows in our use case where connections between the user interface, FL server, and clients must be kept alive until respective FL tasks are carried out, which could be minutes or hours in duration.  
\end{itemize}
\subsection{System Prompt}\label{systemprompt}
The System Prompt is used to explain the LLM's task. It is kept static and supplemented with each intent given to the LLM, both during fine-tuning and inference. The System Prompt used in our work is given as follows 
\begin{quote}
    \textbf{System Prompt:} \textit{Your task is to provide a JSON given an instruction by the user for a federated learning task. The output should be in JSON format. Keys for JSON are algo - Classification/Regression, minibatch - size of minibatch (16), epoch- number of epochs(5), lr - learning rate, scheduler - full/random/round\_robin/latency\_proportional(full), clientFraction- fraction of clients involved in federated learning (1),comRounds- number of communication rounds in federated learning(10), optimizer-pytorch optimizer(Adam), loss(CrossEntropyLoss), compress- No/quantize(No), dtype-img(img),dataset-dataset used for training. Default values for each key are given inside the bracket. Separated by / are possible values for the relevant key. Your task is to create a JSON with the above keys and extract possible values from a given human prompt as values for the JSON. Respond only to the JSON.}
\end{quote}

\end{document}